\documentclass[journal]{IEEEtran}

\usepackage{cite}
\usepackage{amsmath,amssymb,amsfonts}
\usepackage{graphicx, float, booktabs, multirow}
\usepackage{multirow, makecell}
\usepackage{algorithm}
\usepackage{algorithmicx,tikz}
\usepackage{algcompatible}
\usepackage{textcomp,enumitem}
\usepackage{xcolor}
\usepackage[hidelinks]{hyperref}
\def\BibTeX{{\rm B\kern-.05em{\sc i\kern-.025em b}\kern-.08em
    T\kern-.1667em\lower.7ex\hbox{E}\kern-.125emX}}

\usepackage{amsthm}

\begin{document}

\title{%
Operationally Feasible Synthetic Power-Grid Scenarios via Learning the AC-Operable Joint Distribution
}

\author{
Chenhan Xiao,~\IEEEmembership{Student Member,~IEEE,}
Xinyu He,~\IEEEmembership{Student Member,~IEEE,}
Haoran Li,~\IEEEmembership{Student Member,~IEEE,}
Hanghang Tong,~\IEEEmembership{Fellow,~IEEE,}
and~Yang~Weng,~\IEEEmembership{Senior Member,~IEEE.}
\thanks{
C. Xiao, H. Li, and Y. Weng are with the School of Electrical, Computer and Energy Engineering at Arizona State University, USA, e-mail: \{cxiao20, lhaoran, yang.weng\}@asu.edu. 
X. He and H. Tong are with the Siebel School of Computing and Data Science at University of Illinois at Urbana-Champaign, USA, e-mail: \{xhe34, htong\}@illinois.edu.
}
\vspace{-10mm}
}

\maketitle
\begin{abstract}

Synthetic power-grid scenarios are essential for planning, resilience assessment, contingency analysis, and data-driven power-system applications. Recent synthetic grid generation methods have improved structural realism and operational feasibility by incorporating engineering knowledge through post-generation validation, optimization, or physics-aware generation. However, generated scenarios may still exhibit low AC feasibility and robustness, limiting their practical value for downstream power-system studies. This paper proposes a feasibility-aware distribution-learning framework that learns the AC-operable joint distribution of network topology, branch electrical parameters, and time-varying load profiles. Instead of enforcing feasibility after generation, the proposed framework incorporates AC power-flow convergence and operational constraints into hierarchical diffusion-based distribution learning. This enables the generator itself to produce operationally feasible grid scenarios through efficient diffusion sampling. The hierarchical architecture decomposes the high-dimensional generation task into three engineering-motivated stages: topology and bus-attribute generation, branch-parameter generation conditioned on the generated structure, and load-profile generation conditioned on both network structure and electrical characteristics. Experiments on benchmark systems demonstrate that the proposed framework significantly improves operational feasibility and contingency robustness while maintaining strong statistical fidelity and eliminating optimization-based post-processing.
\end{abstract}


\vspace{-0.7em}
\section{Introduction}
\label{sec:intro}

Power-system planning, resilience assessment, contingency screening, and data-driven grid analytics increasingly require large libraries of realistic grid scenarios~\cite{he2026powergrow}. This demand continues to grow as renewable generation, distributed energy resources (DERs), flexible loads, and controllable devices create a much broader range of structural and operating conditions~\cite{li2023distribution}.
Beyond offline studies, online applications such as real-time contingency assessment~\cite{fercEroRTA2021,pnnlWorkflow2024,xiao2023distribution}, corrective topology actions~\cite{li2017rtcaCTS}, and distribution restoration and reconfiguration~\cite{doeFLISR2014} further require rapidly generating large numbers of topology-dependent operating scenarios. 
The value of a synthetic grid generator lies not only in reproducing realistic network structures but also in producing operationally feasible scenarios for downstream power-system analysis.

Generating such scenarios is fundamentally challenging because a power-grid scenario consists of heterogeneous yet tightly coupled variables, including network topology, bus attributes, branch electrical parameters, and time-varying load profiles. Their relationships are governed by the physical hierarchy of power systems: topology determines the admissible electrical structure, electrical parameters determine transfer capability, and operating conditions must remain compatible with both. Consequently, synthetic grid generation is a high-dimensional joint-distribution learning problem constrained by nonlinear AC power-flow equations and engineering operating limits, where feasible scenarios occupy only a small manifold of the overall scenario space~\cite{li2024low}. Directly learning this AC-operable joint distribution while maintaining scalability remains a fundamental challenge.

Existing research has significantly advanced synthetic grid generation from both engineering and machine-learning perspectives. Early methods focused on reproducing structural realism through geographic information systems (GIS), OpenStreetMap data, road networks, and building information~\cite{saha2019framework,dande2024synthetic,baecker2025generation}, statistical matching of network characteristics such as degree distributions, line lengths, and generation-load correlations~\cite{birchfield2016grid,birchfield2016statistical,giacomarra2024generating,gegner2016methodology,schultz2014random}, and graph-mining techniques based on paths, random walks, motifs, and subgraph patterns~\cite{li2026two,yan2021synthetic}. More recently, deep generative models, including GANs, VGAEs, and graph diffusion models, have improved distribution learning by capturing more realistic dependencies among topology and electrical attributes. Representative examples include DeepGDL~\cite{khodayar2019deep}, FeederGAN~\cite{liang2020feedergan}, VGAE-based distribution-grid generation~\cite{abbas2025exploring}, and recent graph diffusion models~\cite{GDSS,GBD}. These approaches improve structural realism and statistical fidelity, but operational feasibility is often treated as a secondary objective or verified after generation rather than being learned as an inherent property of the generative distribution.

Recent physics-aware approaches further improve operational feasibility by incorporating engineering knowledge through staged generation, correction, screening, or optimization-based post-processing~\cite{yan2022active,he2026powergrow}. For example, UG-GAN generates active distribution-system connectivity but subsequently constructs complete test systems through correction and extension procedures~\cite{yan2022active}. Our previous work, PowerGrow, improves feasibility by decomposing structural and dynamic generation and introducing stage-wise constraints~\cite{he2026powergrow}. These methods represent important progress toward operationally realistic synthetic grids. Nevertheless, two fundamental challenges remain. First, operational feasibility is typically enforced through auxiliary constraints, correction, screening, or optimization rather than learned as the primary objective of the joint distribution, so generated scenarios may still exhibit inconsistent AC feasibility and robustness. Second, existing methods do not explicitly learn the AC-operable joint distribution of topology, electrical parameters, and operating conditions, making it difficult to fully capture their nonlinear structural-operational coupling. Motivated by these observations, this paper proposes a feasibility-aware hierarchical diffusion framework that directly learns the AC-operable joint distribution by decomposing generation according to the physical hierarchy of power systems and incorporating AC feasibility into the learning objective rather than treating it solely as post-generation validation.

Directly learning the AC-operable joint distribution is challenging because it couples high-dimensional topology, electrical parameters, and operating conditions with heterogeneous representations and dependencies. These dependencies follow the physical hierarchy of power systems: topology defines the admissible electrical structure, branch parameters determine transfer capability, and operating conditions depend on both. We therefore factorize the joint distribution into hierarchical conditional stages that sequentially generate topology and bus attributes, topology-conditioned branch parameters, and network-conditioned load profiles. Each conditional distribution is modeled using graph beta diffusion~\cite{zhou2023beta,GBD}, which jointly handles sparse connectivity and bounded continuous node and edge attributes.

Hierarchical factorization alone, however, does not ensure that the learned distribution concentrates on AC-operable scenarios. Statistically realistic components may still form infeasible combinations when jointly assembled. We therefore convert AC power-flow convergence and operational constraint violations into offline learning signals that guide the learned distribution toward operable regions, rather than relying on post-generation correction or optimization \cite{yan2022active}. To further control the complexity of long-horizon load generation, a latent temporal representation compresses and reconstructs operating profiles while preserving their dynamics. Together, hierarchical beta-diffusion learning, feasibility-aware distribution refinement, and latent-space load generation enable efficient sampling of complete, operationally feasible scenarios without optimization-based post-processing.

The proposed framework is evaluated on the IEEE 14-bus, 118-bus, 123-bus, and EU 36-bus benchmark systems. Experimental results demonstrate consistently improved AC power-flow convergence, operational feasibility, and contingency robustness while maintaining strong statistical fidelity relative to the baseline methods. Additional ablation studies and downstream applications, including ACOPF and load-stress analysis, further demonstrate the effectiveness and practical utility of the proposed framework.

The rest of this paper is organized as follows. 
Section \ref{sec:system} presents the system modeling for synthetic grid scenario generation. 
Section \ref{sec:method} presents the hierarchical learning framework for the AC-operable joint distribution.
Section \ref{sec:physics} incorporates operational feasibility into the distribution learning process. 
Section \ref{sec:exp} presents numerical evaluations on benchmark power systems. 
Section~\ref{sec:conclusion} concludes the paper.

\vspace{-0.5em}
\section{
System Modeling and Problem Formulation
}
\label{sec:system}

The objective of synthetic power-grid generation is not merely to reproduce realistic network structures or operating profiles independently, but to learn the joint structural-operational distribution from which operationally feasible scenarios can be directly sampled.
Because topology, electrical parameters, and operating conditions jointly determine network transfer capability and AC feasibility, they must be modeled as coupled variables rather than independent generation targets. 
Accordingly, we represent each complete power-grid scenario as
\vspace{-0.3em}
\begin{equation}\label{eq:G}
\mathcal{G}=(X,A,E,D),
\vspace{-0.3em}
\end{equation}
where
\begin{itemize}
    \item $A\in\{0,1\}^{N\times N}$ is the adjacency matrix describing the network topology.
    \item $X\in\mathbb{R}^{N\times d_1}$ contains bus-level attributes, including bus categories (PQ, PV, or slack), generation limits, voltage setpoints, and other operational information.
    \item $E\in\mathbb{R}^{N\times N\times d_2}$ contains branch-level electrical attributes associated with the topology, including resistance, reactance, thermal ratings, and related parameters.
    \item $D\in\mathbb{R}^{N\times T}$ denotes the time-varying load profiles over a horizon of $T$ time steps, represented in this work by bus-level active-power injections.
\end{itemize}

Existing synthetic-grid generation methods often model $X,A,E,D$ through separate or only partially coupled distributions \cite{yan2022active}. 
Such formulations may reproduce the marginal statistics of individual components, but they do not necessarily preserve the nonlinear dependencies that determine AC power-flow operability. 
In particular, a statistically realistic topology may become infeasible when combined with incompatible branch parameters or load profiles.
We therefore formulate synthetic scenario generation as learning the
joint distribution
\vspace{-0.3em}
\begin{equation}\label{eq:joint}
p(X,A,E,D),
\vspace{-0.3em}
\end{equation}
rather than independent distributions such as $p(A)$, $p(E)$, or $p(D)$. 
Eq.~\eqref{eq:joint} defines the central learning target of this work: an AC-operable joint distribution that captures the structural, electrical, and operational dependencies of complete power-grid scenarios. The subsequent methodology is designed to learn this high-dimensional distribution tractably while assigning increased probability mass to operationally feasible regions of the scenario space.
A generated scenario is considered operationally feasible if its AC power-flow equations admit a solution and the resulting operating point satisfies the relevant engineering constraints.
Formally, the grid synthesis problem can then be stated as follows:

\begin{itemize}
    \item {\bf Given}:
    A dataset of real grid scenarios
    $\{\mathcal{G}^{(m)}\}_{m=1}^{M}$,
    where each sample contains jointly observed topology,
    electrical attributes, and load profiles.

    \item {\bf Learn}:
    A generative model that approximates 
    Eq.~\eqref{eq:joint}
    while assigning higher probability mass to operationally
    feasible regions of the scenario space.

    \item {\bf Output}:
    Synthetic scenarios
    $\tilde{\mathcal G}
    =(\tilde X,\tilde A,\tilde E,\tilde D)$
    that preserve the statistical characteristics of real power
    systems while achieving high AC feasibility.
\end{itemize}

\vspace{-1em}
\section{Hierarchical Learning of the AC-Operable Joint Distribution}
\label{sec:method}

Directly learning the joint distribution $p(X,A,E,D)$ in Eq.~\eqref{eq:joint} is challenging because it spans discrete network structures, continuous electrical parameters, and long time-varying load profiles.
Power systems, however, possess a natural physical hierarchy: topology defines the admissible network structure, branch parameters determine its electrical transfer capability, and operating conditions must remain compatible with both. We therefore use this hierarchy to obtain a computationally practical conditional decomposition of the joint distribution.

\vspace{-1em}
\subsection{Hierarchical Generative Factorization}
\label{sec:Hierarchical}
\begin{figure}[h]
    \centering
    \vskip -0.1in
    \includegraphics[width=1\linewidth]{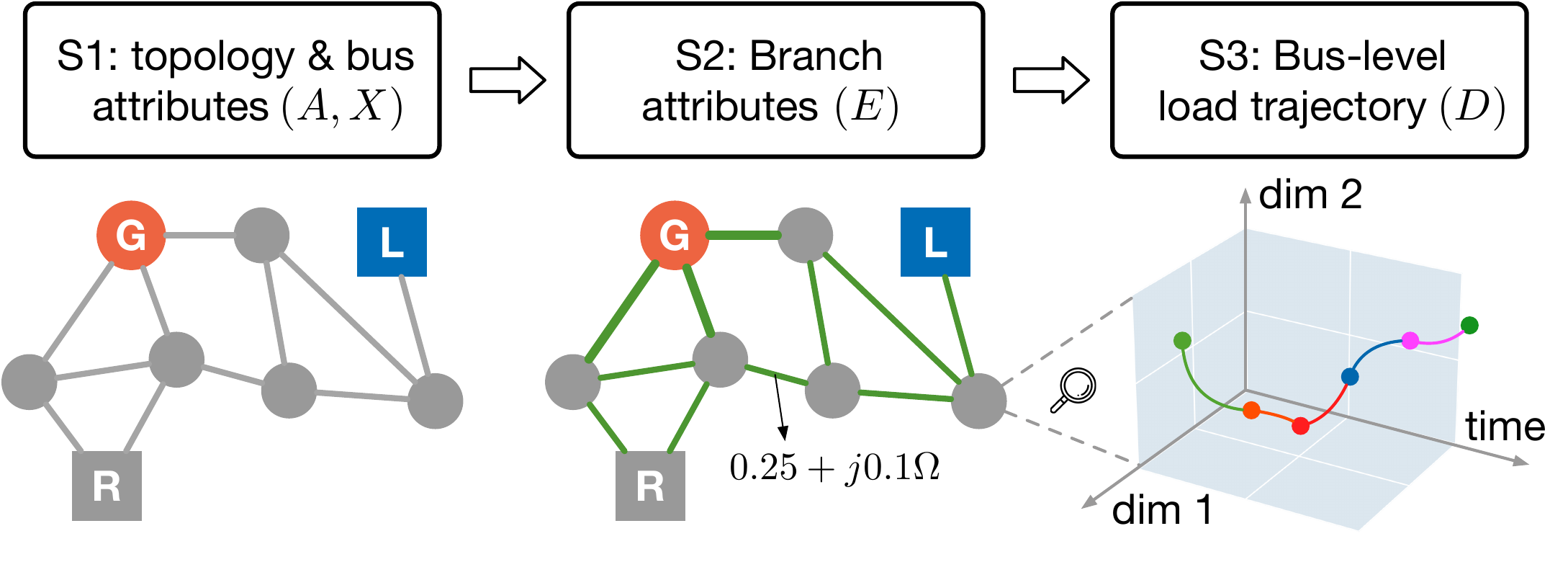}
    \vskip -0.1in
    \caption{Hierarchical factorization of the structural-operational distribution.}
    \label{fig:decompose}
    \vskip -0.1in
\end{figure}

As illustrated in Fig.~\ref{fig:decompose}, the proposed framework learns the joint distribution through three conditional stages. 
Stage~S1 generates the topology and bus attributes $(A,X)$, which define the structural skeleton of the system. 
Stage~S2 generates branch electrical parameters $E$ conditioned on the synthesized structure.
Stage~S3 generates load profiles $D$ conditioned on both the structural and electrical characteristics.

This decomposition yields the hierarchical factorization
\begin{equation}
p(X,A,E,D) = p_{\theta}(A,X)\, p_{\gamma}(E\mid A,X)\, p_{\phi}(D\mid A,X,E).
\label{eq:factor}
\end{equation}
The factorization reflects the physical dependency of power systems: branch parameters are meaningful only after the network structure is specified, while feasible load profiles depend on the topology, branch parameters, and bus attributes.
It also reduces learning complexity because each module models a lower-dimensional conditional distribution rather than the complete joint space, while later stages maintain compatibility with previously generated components.

\vspace{-0.5em}
\subsection{Conditional Learning of Structural and Electrical Distributions}

The first two conditional distributions in Eq.~\eqref{eq:factor} must represent sparse network connectivity, bounded bus attributes, and nonnegative electrical parameters. We therefore adopt graph beta diffusion~\cite{zhou2023beta,GBD,he2026powergrow} as the generative backbone for Stages~S1 and S2. Beta diffusion is suitable for this setting because it supports bounded continuous variables after normalization, accommodates sparse graph representations, and jointly models connectivity and attributed graph information.

A diffusion model consists of a forward process that progressively corrupts observed samples and a learned reverse process that reconstructs samples from noise. Consistent with Eq.~\eqref{eq:factor}, we train two reverse diffusion models:
\begin{equation}
(\hat{A},\hat{X}) = f_{\theta}(\cdot), 
\quad 
\hat{E} = f_{\gamma}(\cdot \mid \hat{A},\hat{X}),
\label{eq:s1s2}
\end{equation}
where $f_{\theta}$ learns the joint structural distribution of topology and bus attributes, and $f_{\gamma}$ learns the conditional distribution of branch parameters on the generated structure.
Both modules are implemented using graph transformers~\cite{DiGress} to capture global connectivity patterns and attribute dependencies. 
We use the standard beta-diffusion training objective (KLUB loss) following~\cite{GBD, zhou2023beta}.

\vspace{-0.7em}
\subsection{Conditional Learning of the Load-Profile Distribution}

The final conditional distribution models time-varying operating conditions given the synthesized network. 
Direct diffusion over raw load trajectories is computationally expensive because the sample dimension grows with both the number of buses and the temporal horizon. Moreover, long trajectories contain substantial temporal redundancy that need not be modeled directly in the diffusion space. We therefore compress the trajectories into a low-dimensional latent representation before conditional generation.

Specifically, an LSTM autoencoder is pretrained to encode and reconstruct
the bus-level load trajectories:
\begin{equation}
H = g_{\mathrm{enc}}(D),\quad \hat{D}=g_{\mathrm{dec}}(H),\quad H\in\mathbb{R}^{N\times d_4},
\label{eq:ae}
\end{equation}
where $d_4 \ll T$.
The encoder--decoder pair is trained using trajectory reconstruction so that $H$ preserves key temporal characteristics, including periodicity, ramps, local fluctuations, and cross-bus correlations.
Stage~S3 then applies beta diffusion in the latent space to learn the
conditional distribution
\begin{equation}
\hat{H} = f_{\phi}(\cdot \mid \hat{A},\hat{X},\hat{E}),
\quad
\hat{D} = g_{\mathrm{dec}}(\hat{H}).
\label{eq:s3}
\end{equation}
This design reduces the computational complexity of diffusion training while retaining the ability to generate long-horizon operating trajectories compatible with the synthesized grid.

Combining the three conditional stages, the complete sampling process is
\begin{equation}
\xrightarrow{\text{S1}} 
(\hat{A},\hat{X}) \ \xrightarrow{\text{S2}} \ \hat{E} \ \xrightarrow{\text{S3}} \ \hat{H} \ \xrightarrow{g_{\mathrm{dec}}} \ \hat{D}.
\end{equation}
The resulting scenario
\(
\tilde{\mathcal G}
=
(\hat X,\hat A,\hat E,\hat D)
\)
constitutes a sample from the learned structural--operational joint distribution. 
The next section introduces feasibility-aware distribution learning to further concentrate this distribution on AC-operable regions.

\vspace{-1em}
\section{
Feasibility-Aware Refinement of the Learned Joint Distribution
}
\label{sec:physics}

Existing synthetic-grid generation methods commonly improve operational feasibility after generation through screening, correction, or optimization. 
In such generate--then--validate approaches \cite{yan2022active}, feasibility serves as an external acceptance criterion and does not guide the learned distribution itself. Consequently, the generator may assign probability mass to operationally infeasible regions.
In contrast, we aim to learn an AC-operable joint distribution from which operationally feasible scenarios can be directly sampled.
Accordingly, we incorporate AC power-flow solvability and operational constraint violations into training as feasibility-guidance signals. 
This transforms feasibility from a post-generation criterion into an inherent property of the learned distribution.

Let $p_{\mathrm{data}}(\mathcal{G})$ denote the empirical distribution of observed grid scenarios and $p_{\alpha}(\mathcal{G})$ the distribution induced by the generative model. Conventional diffusion learning primarily seeks to align $p_{\alpha}(\mathcal{G})$ with $p_{\mathrm{data}}(\mathcal{G})$. However, statistical similarity alone does not ensure that generated combinations of topology, electrical parameters, and operating conditions satisfy AC power-flow physics.
Feasibility-aware distribution learning therefore seeks to preserve statistical fidelity while biasing the learned distribution toward operationally feasible regions. 
This objective is formulated as
\begin{equation}
\max_{p_\alpha}
\;
\mathbb E_{\mathcal G\sim p_\alpha}
\left[
R(\mathcal G)
\right]
\quad
\text{s.t.}
\quad
p_\alpha(\mathcal G)
\approx
p_{\rm data}(\mathcal G),
\label{eq:distribution_shaping}
\end{equation}
where $R(\mathcal G)$ measures the operational quality of a generated scenario. The remainder of this section develops the feasibility metrics and optimization procedure used to realize Eq. \eqref{eq:distribution_shaping}.

\vspace{-0.5em}
\subsection{Operational Score for Distribution Guidance}
\label{sec:feaibility_score}

Realizing Eq.~\eqref{eq:distribution_shaping} requires a learning signal that distinguishes operationally feasible scenarios from infeasible ones. 
We therefore evaluate each generated scenario through AC power-flow simulation and map its solvability and constraint violations to a scalar operational reward. 
This reward allows physical simulation outcomes to guide the learned distribution toward AC-operable regions.

Specifically, we consider two
complementary metrics:
\begin{enumerate}
\item \emph{Power-flow solvability}, indicating whether the AC power-flow equations admit a convergent solution under the synthesized operating conditions. Reactive power is assigned using random power factors uniformly sampled from $[0.85,0.95]$.
\item \emph{Constraint satisfaction}, quantified through a continuous feasibility score that measures the severity of operational constraint violations.
\end{enumerate}

The continuous feasibility score is defined based on the magnitude of operational constraint violations. 
Let the result of an AC power flow simulation be represented by voltage magnitudes $V_i$, phase angles $\theta_i$, and complex power flows $S_{ij} = P_{ij} + j Q_{ij}$ on branch $(i, j)$. 
The constraint violation vector $\mathbf{v}$ aggregates violation magnitudes for key physical constraints:

\begin{itemize}[leftmargin=*]
\item Voltage magnitude limits. For each bus $i$, voltage must remain within its permissible bounds: $V_i^{\min} \leq V_i \leq V_i^{\max}$, yielding the violation
\begin{align}
v_i^{(V)} = \max\Big(0, V_i - V_i^{\max} \Big) + \max\left(0, V_i^{\min} - V_i\right).
\end{align}

\item Branch flow limits. Each transmission line must respect its thermal constraint on apparent power: $|S_{ij}| \leq S_{ij}^{\max}$, yielding the violation
\begin{align}
v_{ij}^{(S)} = \max\left(0, |S_{ij}| - S_{ij}^{\max}\right).
\end{align}

\item Generator capacity limits. Generator outputs must lie within capacity constraints: $P_i^{\min} \leq P_i^{\text{gen}} \leq P_i^{\max}$ and $Q_i^{\min} \leq Q_i^{\text{gen}} \leq Q_i^{\max}$, yielding the violation
\(
v_i^{(P)} = \max(0, P_i^{\text{gen}} - P_i^{\max}) + \max(0, P_i^{\min} - P_i^{\text{gen}})
\) and 
\(
v_i^{(Q)} = \max(0, Q_i^{\text{gen}} - Q_i^{\max}) + \max(0, Q_i^{\min} - Q_i^{\text{gen}}).
\)
\end{itemize}

The complete violation vector is
\begin{equation}
\mathbf{v} = \left[ \{v_i^{(V)}\}_i, \{v_{ij}^{(S)}\}_{(i,j)}, \{v_i^{(P)}\}_i, \{v_i^{(Q)}\}_i \right],
\end{equation}
where generator terms $v^{(P)}$ and $v^{(Q)}$ are included only if generator limits are available in the dataset. 
In experiments, $\mathbf{v}$ is constructed using the raw constraint violation array from the \texttt{runopf} routine in PYPOWER~\cite{pypower}, accessed via \texttt{results["raw"]["g"]}. 
The aggregated violation magnitude is mapped to a continuous feasibility score
\begin{equation}\label{eq:feasibility_score}
    s_{\mathrm{pf}}(\tilde{\mathcal{G}})=\exp\!\left(-\tau\cdot\|\mathbf{v}(\tilde{\mathcal{G}})\|_{1}\right)\in(0,1],
\end{equation}
where the exponential decay yields scores close to 1 for nearly feasible grids and close to 0 for severely infeasible ones, and $\tau$ controls sensitivity.

To jointly account for power-flow solvability and constraint
satisfaction, we define the operational reward
\begin{equation}
R(\tilde{\mathcal G})
=
\lambda_1
\mathbf 1_{\rm converge}(\tilde{\mathcal G})
+
\lambda_2
s_{\rm pf}(\tilde{\mathcal G}),
\label{eq:reward}
\end{equation}
where
$\mathbf 1_{\rm converge}$
is a binary indicator of AC power-flow convergence and $\lambda_1,\lambda_2$ control the relative importance of solvability and feasibility.
The reward in Eq.~\eqref{eq:reward} provides the operational signal that steers the learned distribution toward operationally feasible scenarios as in Eq.~\eqref{eq:distribution_shaping}.

\vspace{-0.5em}
\subsection{Feasibility-Guided Distribution Refinement}
\label{sec:method:training}

Directly optimizing operational reward without first learning the observed scenario distribution may compromise structural, electrical, and temporal fidelity. 
We therefore realize Eq.~\eqref{eq:distribution_shaping} through two complementary stages.
Diffusion pretraining first captures the statistical joint distribution of the training scenarios, after which feasibility-guided refinement shifts probability mass toward its AC-operable regions.

\subsubsection{Diffusion Pretraining}

Let $\mathcal{G}_0 = (A,X,E,H)$ denote a clean training sample and 
$\mathcal{G}_t$ its noisy version under the forward beta-diffusion process. 
The reverse model parameterized by $\alpha$ learns to approximate the true posterior
$q(\mathcal{G}_{t-1}|\mathcal{G}_t,\mathcal{G}_0)$.

Following~\cite{GBD, zhou2023beta}, diffusion pretraining
minimizes
\begin{equation}
\mathcal{L}_{\text{diff}}
=
\mathbb{E}_{t,\mathcal{G}_0,\mathcal{G}_t}
\left[
\mathrm{KL}
\big(
p_\alpha(\mathcal{G}_{t-1}|\mathcal{G}_t)
\;\|\;
q(\mathcal{G}_{t-1}|\mathcal{G}_t,\mathcal{G}_0)
\big)
\right].
\end{equation}

This objective is applied to the three hierarchical components defined in Section~\ref{sec:Hierarchical}, ensuring that the model captures the statistical structure of topology, branch parameters, and latent load embeddings.

\subsubsection{Feasibility-Based Refinement}
While diffusion pretraining approximates the empirical data
distribution, it does not explicitly distinguish between
operationally feasible and infeasible scenarios. We therefore
refine the pretrained model using the operational reward in Eq.
\eqref{eq:reward}.

Let
$p_\alpha(\tilde{\mathcal G})$
denote the distribution induced by the current diffusion
model. Consistent with the feasibility-guidance objective, model parameters are updated by solving
\(
\max_{\alpha}
\mathbb E_{\tilde{\mathcal G}\sim p_\alpha}
R(\tilde{\mathcal G}).
\)
Using the score-function estimator, the gradient is
approximated by
\(
\nabla_\alpha
\mathbb E[R]
\approx
\mathbb E_{\tilde{\mathcal G}\sim p_\alpha}
R(\tilde{\mathcal G})
\nabla_\alpha
\log p_\alpha(\tilde{\mathcal G}).
\)
In practice, mini-batch samples are drawn from the current model, evaluated using AC simulation, and used to update the model parameters through this reward-weighted gradient.
Combining diffusion learning and feasibility-guided distribution refinement yields
\begin{equation}\label{eq:total_loss}
\mathcal{L}_{\text{total}}
=
\mathcal{L}_{\text{diff}}
-
\beta
\,
\mathbb{E}_{\tilde{\mathcal{G}}\sim p_\alpha}
R(\tilde{\mathcal{G}}),
\end{equation}
where $\beta$ controls the strength of feasibility guidance. 
In Eq.~\eqref{eq:total_loss}, the diffusion term preserves fidelity to the empirical structural--operational distribution, corresponding to the distributional constraint in Eq.~\eqref{eq:distribution_shaping}, while the reward term increases probability mass in AC-operable regions. The resulting model therefore learns an operationally guided yet statistically realistic joint distribution. During inference, complete scenarios are obtained through standard diffusion sampling without power-flow-based correction or optimization.

\vspace{-0.5em}
\section{Numerical Studies
}\label{sec:exp}

The numerical studies evaluate whether the proposed framework
(1) preserves the structural, electrical, and temporal characteristics of the reference systems, 
(2) generates grid scenarios that admit feasible AC power-flow solutions,
(3) benefits from hierarchical factorization and feasibility-aware refinement, 
and (4) supports direct use in representative planning and security-oriented applications.

\vspace{-0.5em}
\subsection{Experiment Setup}
\label{sec:experiment_setup}

We evaluate the framework on representative transmission and distribution benchmarks widely used in power-system studies:
(1) IEEE 14-bus transmission system, a compact meshed network with five generators,
(2) European (EU) 36-bus urban distribution grid~\cite{mateo2018european}, a medium-voltage network with a single slack generator and predominantly radial topology,
(3) IEEE 118-bus transmission system and (4) IEEE 123-bus distribution feeder. The latter two provide larger-scale benchmarks for assessing scalability.
These four systems collectively cover meshed transmission grids and radial distribution feeders across multiple system scales. Extremely large transmission systems (e.g., thousands of buses) are not considered here because they rarely form uniformly dense graphs. Instead, they exhibit modular structures induced by substations and regional planning boundaries. Such articulation points partition the network into weakly coupled subnetworks, allowing large grids to be synthesized through smaller subnetwork generation followed by sparse interconnection modeling.

For each base system, we construct a scenario library of 2{,}000 topology variants, which form the training dataset for that system. 
Each variant is generated using (1) degree-preserving edge swaps and local subgraph perturbations implemented via a random-walk procedure and (2) mild perturbations of branch electrical parameters (e.g., impedance variations within realistic bounds).
These perturbations emulate practical grid changes such as switching actions, line-parameter uncertainty, and operational variability, rather than arbitrary graph randomization.

The above random-walk process intentionally produces a mixture of operable and non-operable configurations, including islanding and topologies that violate operational constraints under realistic loading conditions. This prevents the training dataset from being restricted to trivially feasible networks and reflects the practical difficulty of preserving feasibility under heuristic structural modifications.
For each grid variant, AC power flow simulations are performed using \texttt{PYPOWER}~\cite{pypower} with anonymized hourly load trajectories from Duquesne Light Company (Pittsburgh, USA). To introduce operating variability and avoid overfitting to a single profile, both load and generation levels are randomly scaled. Measurement noise with $2\%$ standard deviation is injected to emulate typical SCADA accuracy levels.

\vspace{-0.5em}
\subsection{Baselines and Evaluation Metrics}
We compare the proposed approach with representative synthetic grid generation methods from both the power-systems and machine-learning literature.

\begin{itemize}[leftmargin=*]
\item \textbf{Power-system grid synthesis methods.}
We consider three representative domain-specific approaches.
The first is the statistics-alignment construction method (StatAlign)~\cite{birchfield2016grid}, which synthesizes grids by matching global structural statistics (e.g., degree distribution, clustering coefficient, and generation-load ratios) and then assigning electrical parameters.
The second is a power-system deep generative model (GNN-Gen)~\cite{liu2023gnn}, a GNN-based generator built on graph variational autoencoders that learns feature distributions of real grids.
The third is PowerGrow~\cite{he2026powergrow}, a hierarchical
generation framework that also jointly synthesizes grid topology
and load dynamics.
Unlike the proposed method, PowerGrow does not explicitly incorporate AC
feasibility as a feasibility-guidance learning signal.

\item \textbf{General graph diffusion models.}
To benchmark against modern attributed graph generators from the machine-learning literature, we include representative diffusion models including  GDSS~\cite{GDSS}, GruM~\cite{GruM}, and GBD~\cite{GBD}.
These methods are designed for generic graph generation with continuous attributes and are not tailored to power-system operational constraints.
For fairness, we incorporate the same operational feasibility guidance mechanism during training.

\item \textbf{Random-walk perturbation baseline.}
We also include the topology perturbation procedure used to construct the training dataset as a non-learning baseline, representing purely structural randomization without learned generative modeling.
\end{itemize}

All baselines are adapted to generate topology instances with associated load profiles. For methods (StatAlign and GNN-Gen) that do not natively produce load trajectories end-to-end, we synthesize plausible load profiles using optimization-based load construction techniques \cite{yan2022active}.

\begin{table*}[t]
\centering
\caption{Operational validation across four test systems. Conv.: AC power-flow convergence rate (\%); 
Feas.: feasibility score in $[0,1]$; 
$N$--1: single-line-outage survivability rate (\%). Higher value is better.}
\label{tab:op_metrics_all_systems}
\setlength{\tabcolsep}{4.5pt}
\renewcommand{\arraystretch}{1.08}
\begin{tabular}{lccc ccc ccc ccc}
\toprule
\multirow{2}{*}{\textbf{Method}} 
& \multicolumn{3}{c}{\textbf{IEEE 14-bus}} 
& \multicolumn{3}{c}{\textbf{EU 36-bus}} 
& \multicolumn{3}{c}{\textbf{IEEE 118-bus}} 
& \multicolumn{3}{c}{\textbf{IEEE 123-bus}} \\
\cmidrule(lr){2-4} \cmidrule(lr){5-7} \cmidrule(lr){8-10} \cmidrule(lr){11-13}
& Conv. & Feas. & $N$--1 
& Conv. & Feas. & $N$--1 
& Conv. & Feas. & $N$--1 
& Conv. & Feas. & $N$--1 \\
\midrule

Ours                                    & 98.6 & 0.934 & 76.8 & 94.5 & 0.912 & -- & 92.8 & 0.876 & 68.5 & 100 & 0.581 & -- \\
PowerGrow \cite{he2026powergrow}        & 93.2 & 0.806 & 65.4 & 86.5 & 0.774 & -- & 85.4 & 0.721 & 55.7 & 100 & 0.529 & -- \\
Random-walk                             & 88.5 & 0.662 & 48.9 & 81.3 & 0.696 & -- & 78.6 & 0.604 & 41.3 & 100 & 0.315 & -- \\
StatAlign \cite{birchfield2016grid}     & 84.8 & 0.247 & 36.2 & 81.9 & 0.238 & -- & 75.8 & 0.189 & 30.6 & 96.4 & 0.201 & -- \\
GNN-Gen \cite{liu2023gnn}               & 82.6 & 0.382 & 40.5 & 82.0 & 0.353 & -- & 76.5 & 0.311 & 34.8 & 97.2 & 0.244 & -- \\
GDSS \cite{GDSS}                        & 87.1 & 0.428 & 46.7 & 84.2 & 0.326 & -- & 80.2 & 0.354 & 39.5 & 98.6 & 0.276 & -- \\
GruM \cite{GruM}                        & 88.4 & 0.531 & 52.1 & 84.5 & 0.562 & -- & 81.1 & 0.497 & 45.2 & 99.1 & 0.335 & -- \\
GBD \cite{GBD}                          & 90.2 & 0.487 & 50.6 & 85.9 & 0.453 & -- & 82.4 & 0.421 & 43.7 & 99.3 & 0.392 & -- \\
\bottomrule
\end{tabular}
\end{table*}

To assess the quality of generated samples, we adopt the following evaluation metrics.

\begin{itemize}[leftmargin=*]
\item {\bf Power-flow solvability.} We report the \emph{convergence rate}, defined as the fraction of generated samples whose AC power flow converges under the synthesized load profiles. Reactive demand is assigned using random power factors uniformly sampled from $[0.85, 0.95]$.

\item {\bf Feasibility score}. Beyond binary convergence, we evaluate operational validity using the \emph{feasibility score} (Section~\ref{sec:feaibility_score}), which aggregates voltage, thermal, and generator constraint violations into a continuous metric. Owing to exponential normalization, higher values indicate operation closer to the feasible region.

\item {\bf $N\!-\!1$ reliability}. For each convergent transmission-grid sample, we remove each branch and re-solve the AC power flow. The \emph{$N\!-\!1$ convergence rate} is reported as the fraction of outages that remain solvable, serving as a proxy for structural redundancy and reliability.

\item {\bf Statistical fidelity}. Statistical alignment is evaluated using Maximum Mean Discrepancy (MMD) between generated and reference distributions over structural, temporal, and attribute characteristics, including (1) graph degree distribution (Deg.), (2) clustering coefficient (Clus.), (3) orbit counts (Orbit.), (4) spectral properties (Spec.), (5) load time-series statistics (Time.), and (6) branch and node attributes (Attr.). Lower MMD indicates better alignment.
\end{itemize}

All experiments are conducted on the ASU Sol high-performance computing cluster,
using four NVIDIA A100 GPUs (80\,GB memory each). 
To stabilize diffusion training, branch electrical parameters and active-power trajectories are normalized to $[0,1]$, and rescaled to physical units after generation.
The diffusion model is trained using the Adam optimizer with learning rate $10^{-4}$, batch size 50, and 1{,}000 diffusion steps. 
The LSTM autoencoder uses latent dimension $d_4=64$.
The model is pretrained for 600 epochs and feasibility-refined for 5{,}400 epochs.
For feasibility-aware refinement, we set $\tau=10^{-5}$ in Eq.~\eqref{eq:feasibility_score}, $\lambda_1=\lambda_2=0.5$ in Eq.~\eqref{eq:reward}, and $\beta=0.01$ in Eq.~\eqref{eq:total_loss}.

\vspace{-0.8em}
\subsection{Quantitative Comparison Across Methods}
\label{sec:exp:compare-baselines}

Table~\ref{tab:op_metrics_all_systems} summarizes the operational performance of scenarios sampled from the learned distributions across all benchmark systems. For each method, 100 scenarios are generated, and the metrics are averaged. The proposed method consistently outperforms the baselines across all operational metrics, demonstrating the benefit of combining feasibility-aware distribution learning with the physically motivated hierarchical factorization. Despite receiving the same feasibility guidance, the general graph diffusion baselines remain less effective because their monolithic formulations do not explicitly preserve the conditional dependencies among topology, electrical parameters, and operating conditions.

The proposed method achieves power-flow convergence rates of 98.6\% and 94.5\% on the IEEE 14-bus and EU 36-bus systems, respectively, and maintains strong performance on the larger IEEE 118-bus and IEEE 123-bus networks. This indicates that most generated scenarios admit physically meaningful AC power-flow solutions suitable for downstream studies. The feasibility scores further support this trend: the proposed method achieves the highest score on every benchmark. Under the $N\!-\!1$ metric, reported for the transmission systems, the proposed method also maintains strong contingency robustness.

\begin{figure}[H]
    \centering
    \vskip -0.15in
    \includegraphics[width=1\linewidth]{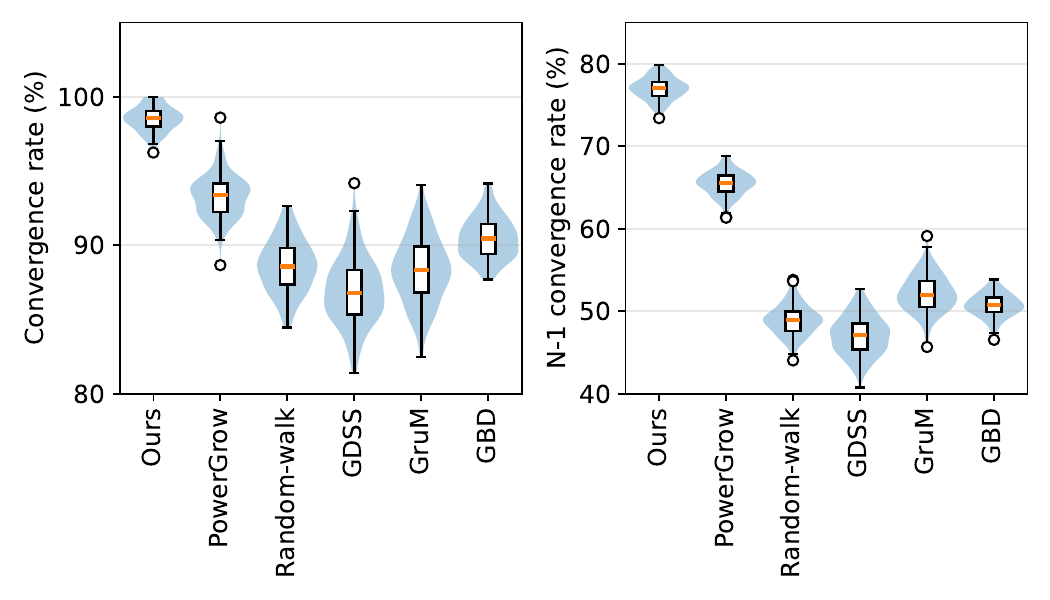}
    \vskip -0.1in
    \caption{Violin plots of distribution of operational metrics for generated IEEE 14-bus scenarios.}
    \label{fig:violin}
    \vskip -0.1in
\end{figure}

Beyond the averaged metrics in Table~\ref{tab:op_metrics_all_systems}, we examine their variability across 20 independent trials, each comprising 100 generated scenarios.
Fig.~\ref{fig:violin} reports the trial-level distributions of the AC power-flow convergence rate and the $N\!-\!1$ convergence rate on the IEEE 14-bus system for methods that jointly generate topology and load trajectories. The proposed method achieves the highest and most concentrated distributions for both metrics, with AC convergence rates clustered near $98$--$99\%$ and $N\!-\!1$ convergence rates centered in the upper 70\% range.
PowerGrow also performs strongly but exhibits a lower median and wider spread, especially under the $N\!-\!1$ test. In contrast, random-walk samples show substantially lower robustness despite moderate nominal convergence, reflecting the fact that local edge perturbations can preserve coarse topology statistics while weakening contingency margins. Among diffusion-based baselines, GDSS, GruM, and GBD achieve reasonable nominal convergence but noticeably weaker $N\!-\!1$ robustness. GruM and GBD have slightly higher robustness medians than GDSS, but all remain well below the proposed method. 
These results demonstrate that the proposed combination of feasibility guidance and hierarchical factorization improves not only average performance but also stability across experimental trials.
Such consistency is important for planning and contingency analysis, where even a small number of infeasible scenarios can distort risk assessment. 

To further evaluate statistical fidelity, Table~\ref{tab:exp:mmd} reports MMD scores on the EU 36-bus system across structural, temporal, and attribute distributions. 
As expected, the random-walk baseline achieves extremely low MMD on several structural metrics because generated samples remain close to the original topology through local edge rewiring. Similarly, StatAlign obtains near-perfect agreement in degree, clustering, and orbit statistics by explicitly matching selected graph descriptors. However, these methods are designed to preserve statistical properties rather than operational characteristics, and their strong structural alignment does not translate directly into improved feasibility or grid performance, as evidenced by the operational results in Table~\ref{tab:op_metrics_all_systems}.

\begin{table}[h]
    \centering
    \vskip -0.1in
    \caption{Comparison of MMD metrics on the EU 36-bus system. All reported scores are scaled by $10^{-2}$.}
    \vskip -0.05in
    \begin{tabular}{ccccccc}
        \toprule
        Methods & Deg.  & Clus.  & Orbit  & Spec.  & Time.  & Attr.  \\
        \midrule
        Ours                   & 0.24 & 0.10 & 0.02 & 0.11 & 0.07 & 0.13  \\
        PowerGrow \cite{he2026powergrow}     & 0.22 & 0.09 & 0.02 & 0.10 & 0.11 & 0.14  \\
        Random-walk            & 0.00  & 0.08 & 0.03 & 0.09 & 0.18 & 0.001  \\
        StatAlign \cite{birchfield2016grid} & 0.047 & 0.00 & 0.001 & 0.46 & 0.42 & 0.001 \\
        GNN-Gen \cite{liu2023gnn} & 0.67 & 0.86 & 0.73 & 2.21 & 0.35 & 1.13 \\
        GDSS \cite{GDSS}       & 0.26 & 0.17 & 0.04 & 0.08 & 0.16 & 0.14  \\
        
        GruM \cite{GruM}       & 0.23 & 0.26 & 0.04 & 0.19 & 0.22 & 0.23  \\

        GBD \cite{GBD}         & 0.19 & 0.07 & 0.01 & 0.09 & 0.14 & 0.11  \\
        \bottomrule
    \end{tabular}
    \label{tab:exp:mmd}
\vskip -0.05in
\end{table}

Among learning-based baselines, GBD and GDSS provide the strongest overall structural fidelity, achieving low discrepancies across degree, clustering, orbit, and spectral statistics. In contrast, GNN-Gen exhibits substantially larger MMD values across most categories, suggesting difficulty in accurately reproducing the distribution of realistic grid topologies under the limited training regime considered here. For temporal statistics, methods that rely on post-hoc load-profile construction, such as StatAlign and GNN-Gen, exhibit noticeably larger discrepancies than diffusion-based approaches that model topology and load characteristics more coherently.
For the proposed method, although its structural MMD scores are slightly higher than those of the purely distribution-matching GBD baseline, the differences remain small, particularly in orbit and spectral statistics. More importantly, these small differences in statistical fidelity are accompanied by substantially higher convergence and feasibility as shown in Table~\ref{tab:op_metrics_all_systems}.

Generation speed is another key consideration in generative modeling, as excessive sampling time limits scalability and iterative refinement. Table~\ref{tab:runtime} reports the average end-to-end generation time per scenario, including topology, branch-attribute, and load-trajectory synthesis, on the EU 36-bus system. The proposed method generates a complete scenario in 0.637 s/sample, which is only moderately slower than PowerGrow (0.625 s/sample) and remains comparable to diffusion-based baselines such as GBD and GruM. Although our framework decomposes generation into three diffusion stages, each submodel operates on a smaller and more structured subproblem than the monolithic GBD baseline, so the overall online sampling overhead remains limited.

In contrast, StatAlign and GNN-Gen require optimization-based load
construction after topology generation, resulting in substantially
longer runtimes of 31--36~s/sample. The proposed method directly
samples grid scenario from the
learned model, where the expensive feasibility evaluation and reward-guided refinement are
performed only during offline training. 
This offline cost is acceptable
because the trained model supports repeated large-scale studies
requiring fast scenario generation.
Together with the high
convergence rates in Table~\ref{tab:op_metrics_all_systems}, these
results demonstrate high-yield direct scenario generation without
optimization-based post-processing. 

\begin{table}[h]
\centering
\vskip -0.1in
\caption{Average sampling time per scenario (topology, branch attributes, and load trajectory) on the EU 36-bus system.}
\label{tab:runtime}
\vskip -0.1in
\setlength{\tabcolsep}{6pt}
\renewcommand{\arraystretch}{1.1}
\begin{tabular}{l c}
\toprule
\textbf{Method} & \textbf{Time per sample (sec)} \\
\midrule
Ours & 0.637 \\
PowerGrow \cite{he2026powergrow} & 0.625 \\
StatAlign \cite{birchfield2016grid} + OPT-based \cite{yan2022active} & 36.247 \\
GNN-Gen \cite{liu2023gnn} + OPT-based \cite{yan2022active} & 31.389 \\
GDSS \cite{GDSS} & 2.620 \\
GruM \cite{GruM} & 0.660 \\
GBD \cite{GBD} & 0.501 \\
\bottomrule
\end{tabular}
\end{table}

\vspace{-1em}
\subsection{Qualitative Demonstrations of Generated Scenarios}
\label{sec:exp:visualize}

\begin{figure}[H]
    \centering
    \vskip -0.15in
    \includegraphics[width=1\linewidth]{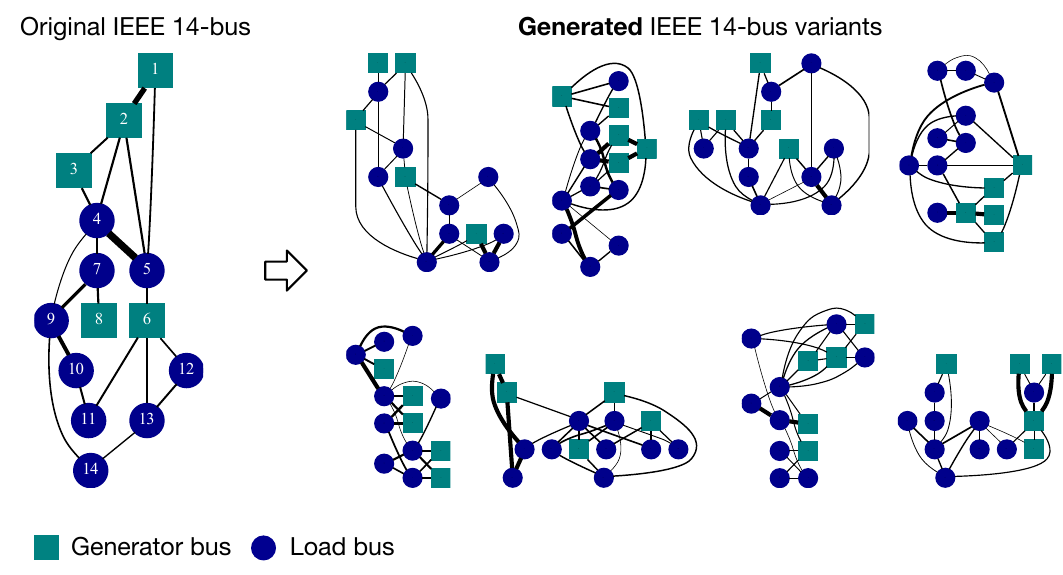}
    \vskip -0.1in
    \caption{Original IEEE 14-bus transmission network (left) and example generated topology variants (right). 
    Branch thickness is inversely proportional to the normalized magnitude of branch reactance.}
    \label{fig:visualize_14}
    \vskip -0.1in
\end{figure}

\begin{figure}[h]
    \centering
    \vskip -0.1in
    \includegraphics[width=1\linewidth]{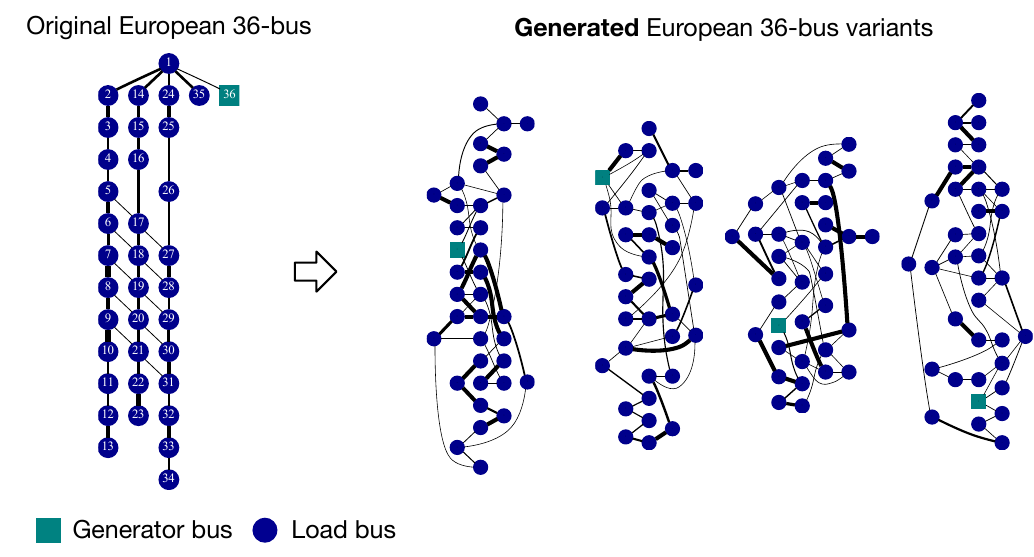}
    \vskip -0.1in
    \caption{Original EU 36-bus medium-voltage urban distribution network (left) and example generated topology variants (right).}
    \label{fig:visualize_36}
    \vskip -0.1in
\end{figure}

Beyond quantitative metrics, we present visual comparisons of synthesized topologies and load profiles to illustrate the realism and diversity of the generated scenarios. 
Figs.~\ref{fig:visualize_14} and~\ref{fig:visualize_36} compare the original benchmark networks with representative generated variants for the IEEE 14-bus transmission system and the EU 36-bus distribution network.

For the IEEE 14-bus system (Fig.~\ref{fig:visualize_14}), the generated grids preserve key structural characteristics of transmission grids, including generator-load separation, a meshed backbone, and multi-path connectivity, while exhibiting nontrivial variations in local connectivity patterns. 
Branch thickness, proportional to $1/|x|$, reflects electrical coupling strength and highlights major transmission corridors. 
For the EU 36-bus system (Fig.~\ref{fig:visualize_36}), which represents a predominantly radial medium-voltage distribution network, the generated graphs maintain radial dominance and feeder-like structures while avoiding disconnected components. 
The synthesized variants display realistic sub-feeder branching and occasional localized meshing that resemble practical distribution layouts. 
These results suggest that the proposed model adapts naturally to fundamentally different structures (meshed transmission networks versus radial distribution systems) without requiring architecture-specific modifications.

\begin{figure}[h]
    \centering
    \vskip -0.1in
    \includegraphics[width=1\linewidth]{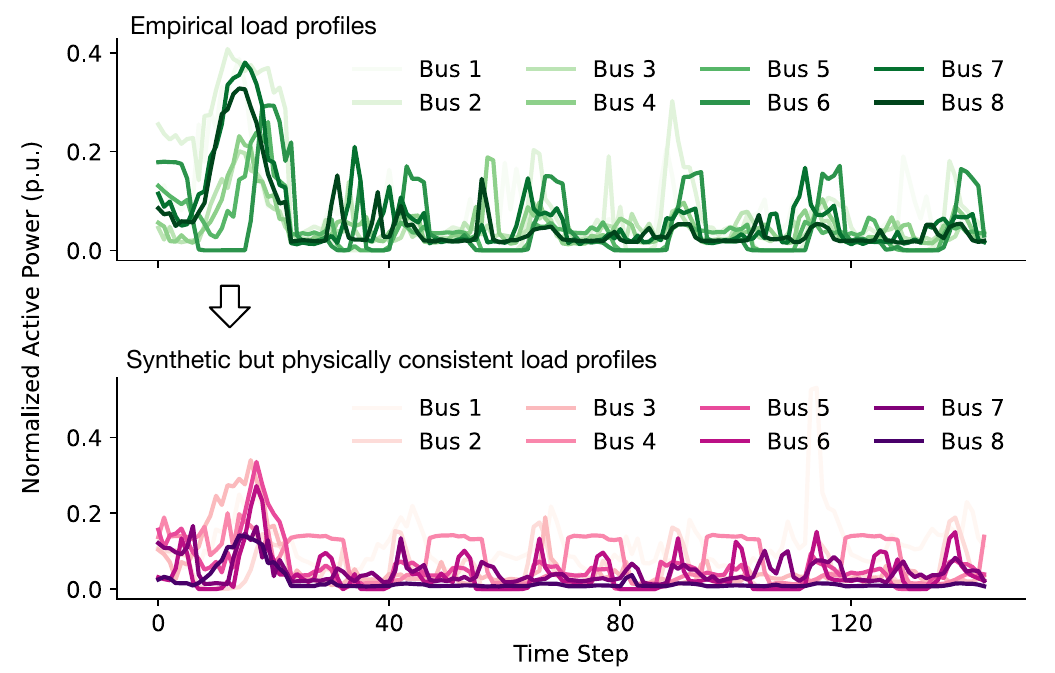}
    \vskip -0.1in
    \caption{Top: normalized historical bus-level active power trajectories used for training. Bottom: load trajectories generated by the proposed model. Results are shown for the EU 36-bus medium-voltage urban network.}
    \label{fig:load-generation}
    \vskip -0.1in
\end{figure}

For the load dynamics, Fig.~\ref{fig:load-generation} compares the hourly training trajectories (top) with generated samples (bottom) for representative buses in the 36-bus system. 
The synthesized trajectories reproduce key temporal patterns observed in the training data, including daily periodicity, smooth baseline evolution, and occasional localized fluctuations. 
Importantly, the generated loads remain within physically plausible ranges and exhibit correlated variability across buses, indicating that the model captures correlated spatiotemporal structure rather than independent noise. 
The similarity in amplitude range, variability scale, and temporal smoothness suggests that the diffusion process successfully learns the statistical and physical regularities governing load dynamics.

\vspace{-0.5em}
\subsection{Evaluation of Impact of Feasibility Guidance}
\label{sec:exp-mechanism}

\begin{figure}[h]
    \centering
    \vskip -0.1in
    \includegraphics[width=1\linewidth]{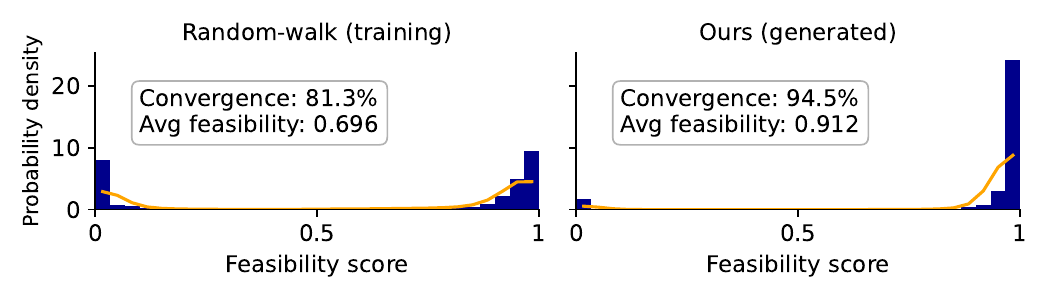}
    \vskip -0.1in
    \caption{Distribution of feasibility scores on the 36-bus system.
Left: random-walk perturbed training samples. Right: samples generated by our model.}
    \label{fig:histogram}
    \vskip -0.1in
\end{figure}

To examine how feasibility guidance reshapes the generative
distribution, we first illustrate how it biases generation toward operationally feasible regions. Fig.~\ref{fig:histogram} compares the feasibility score distributions of the random-walk training data and the generated samples.
The random-walk perturbation process produces a mixed-quality dataset containing a substantial fraction of infeasible samples. In particular, only 81.3\% of the perturbed grids achieve power-flow convergence, with an average feasibility score of 0.696. The resulting distribution is widely dispersed, with noticeable probability mass in low-feasibility regions.

Despite being trained on this imperfect dataset, the proposed model learns to reshape the underlying distribution toward its operationally valid subset. The generated samples achieve a 94.5\% convergence rate and an average feasibility score of 0.912, with the distribution tightly concentrated near the high-feasibility end of the score range. This improvement reflects not only an increase in the mean feasibility score but also a systematic redistribution of probability mass away from infeasible regions.
These results indicate that feasibility-aware guidance actively encourages the diffusion model to generate AC-feasible scenarios under realistic load trajectories.

\begin{figure}[h]
    \centering
    \vskip -0.1in
    \includegraphics[width=1\linewidth]{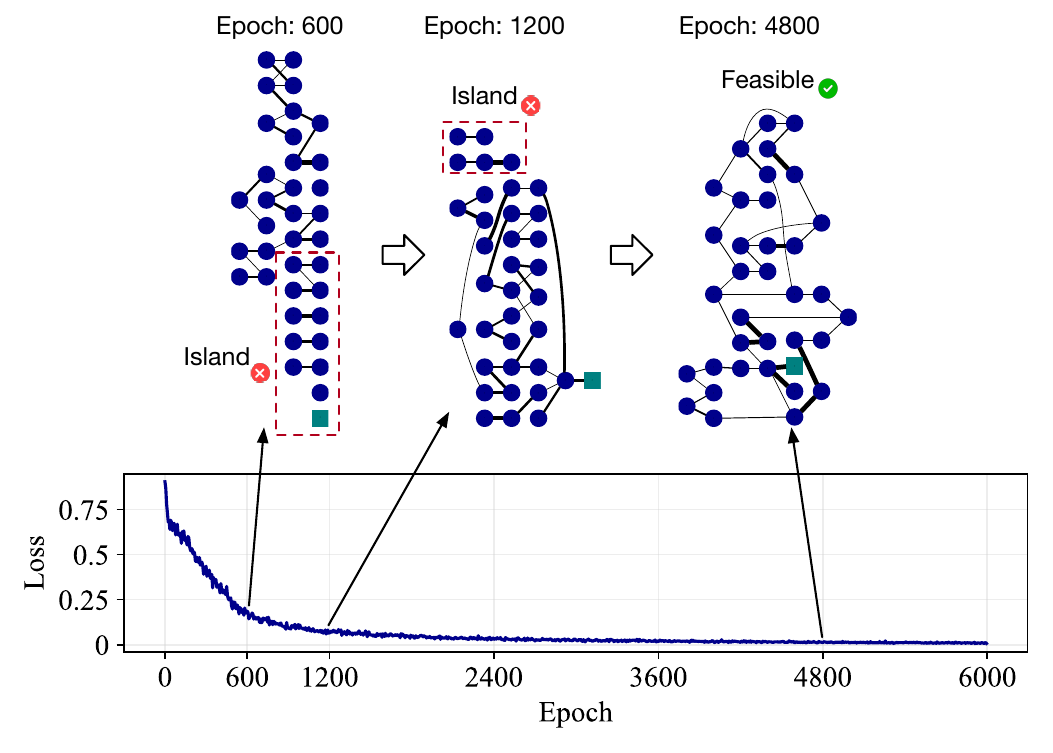}
    \vskip -0.15in
    \caption{Evolution of generated topology across training epochs in the EU 36-bus dataset. The training loss trajectory is shown in the lower panel.}
    \vskip -0.1in
    \label{fig:tra-topology}
\end{figure}

To further understand how feasibility guidance reshapes the learned distribution, we examine the evolution of generated grids and load trajectories throughout training.
Fig.~\ref{fig:tra-topology} visualizes representative topologies sampled at different training epochs for the EU 36-bus system. Early-stage generations (e.g., epochs 600–1200) frequently exhibit disconnected components (``islands'') and irregular structural patterns. Such configurations are physically undesirable because they correspond to infeasible or weakly connected grids that fail AC power-flow validation.
As training progresses, the diffusion model gradually learns structural regularities. By later epochs (e.g., around epoch 4800), the generated graphs become more coherent, with improved connectivity and fewer structural anomalies. Importantly, this structural refinement emerges without explicitly enforcing connectivity constraints during generation. Instead, the integration of feasibility-aware guidance into the diffusion objective biases the sampling process toward topologies that satisfy AC operational requirements.

\begin{figure}[h]
    \centering
    \vskip -0.1in
    \includegraphics[width=1\linewidth]{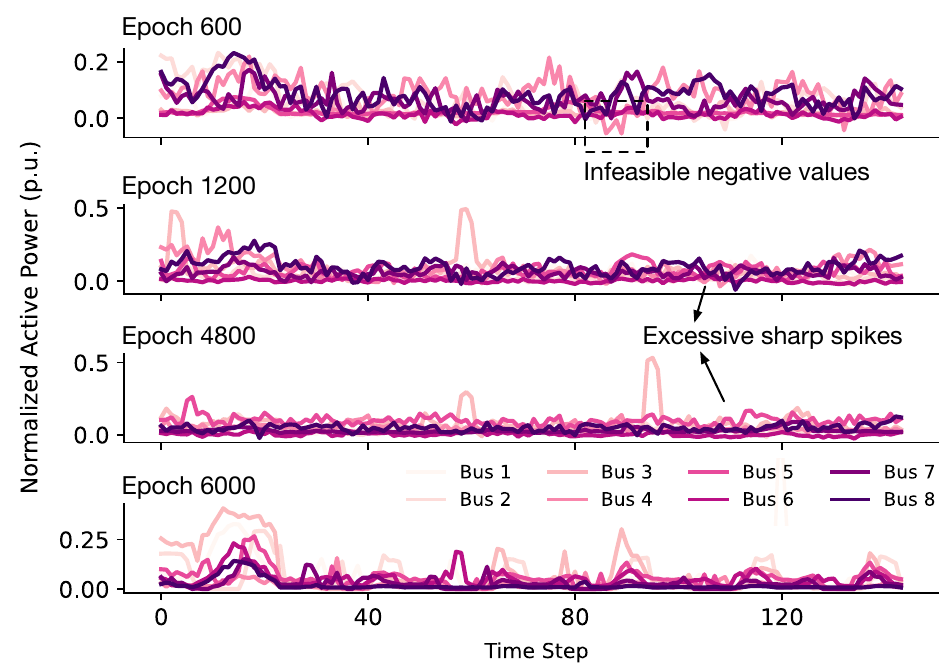}
    \vskip -0.1in
    \caption{Evolution of generated bus-level load trajectories during training in the EU 36-bus network.}
    \label{fig:tra-timeseries}
    \vskip -0.05in
\end{figure}

In addition to topology evolution, we examine how the generated time-series load trajectories improve during training. Fig.~\ref{fig:tra-timeseries} presents representative normalized active power sequences (in per unit) sampled at different epochs. At early stages (e.g., epoch 600), the generated profiles display instability, including infeasible negative values and abrupt high-frequency spikes. Although some periodic patterns appear due to LSTM-autoencoder pretraining, the temporal dynamics remain unrealistic and poorly aligned with empirical load behavior.
By intermediate epochs (e.g., epochs 1200–4800), the trajectories begin to exhibit smoother patterns with reduced spurious oscillations. The model increasingly captures realistic temporal structure while suppressing physically implausible fluctuations. At later epochs (e.g., epoch 6000), the generated load profiles demonstrate stable, smooth, and diverse trajectories that resemble realistic residential or commercial demand patterns. The disappearance of infeasible values and excessive spikes reflects the model’s improved alignment with operational constraints under realistic load dynamics.

Finally, we conduct an ablation study on the IEEE 14-bus system to isolate the contributions of feasibility guidance and hierarchical factorization. Specifically, we compare
(1) the full model with both feasibility guidance and hierarchical factorization,
(2) a variant without feasibility guidance in the training phase, 
and
(3) a single-diffusion variant that jointly generates topology, branch attributes, and load trajectories without hierarchical decomposition.
As shown in Table~\ref{tab:ablation}, removing feasibility guidance significantly degrades performance across all metrics, reducing convergence rate from 98.6\% to 86.5\%, feasibility score from 0.934 to 0.812, and $N$–1 convergence from 76.8\% to 54.7\%. This confirms that feasibility guidance is the primary driver steering the learned distribution toward operationally valid regions.
Replacing the hierarchical design with a single diffusion model also degrades performance, as jointly generating topology, attributes, and load trajectories within a single process increases task complexity and makes optimization more difficult, thereby reducing training stability and solution quality.

\begin{table}[h]
\centering
\caption{Ablation study on the IEEE 14-bus system.}
\label{tab:ablation}
\setlength{\tabcolsep}{6pt}
\begin{tabular}{lccc}
\toprule
Model Variant & Conv. (\%)  & Feas. Score  & $N$--1 (\%)  \\
\midrule
Full Model & 98.6 & 0.934 & 76.8 \\
w/o Feasibility Guidance & 86.5 & 0.812 & 54.7 \\
w/o Hierarchical Factorization & 91.3 & 0.874 & 60.8 \\
\bottomrule
\end{tabular}
\end{table}

\subsection{Assessment in Planning and Security-Oriented Applications}

While the preceding sections establish the statistical fidelity and operational feasibility of the generated scenarios, the ultimate value of a scenario generator lies in its usefulness for downstream power-system studies. To evaluate whether the synthesized grid scenarios are meaningful for practical analysis, we examine their performance in representative planning and security-oriented applications. Specifically, we consider (i) AC optimal power flow (ACOPF) to evaluate economic dispatch behavior, and (ii) progressive load stress to assess system robustness under increasing demand.

\subsubsection{AC Optimal Power Flow Analysis}

\begin{figure}[h]
    \centering
    \vskip -0.1in
    \includegraphics[width=1\linewidth]{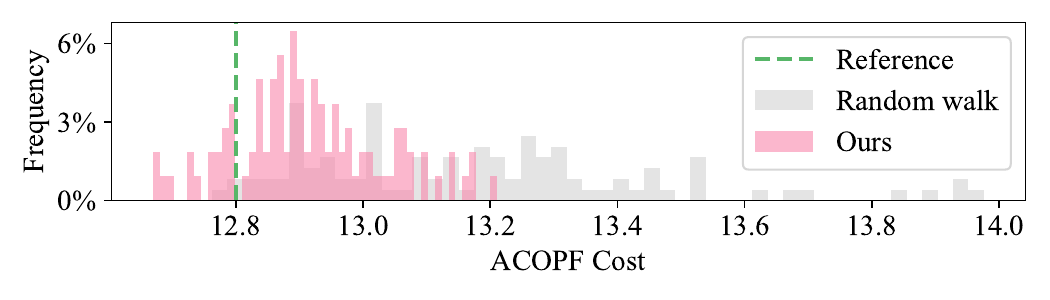}
    \vskip -0.15in
    \caption{Histogram of ACOPF costs for grids generated by the proposed model and the random-walk procedure on the EU 36-bus system.}
    \label{fig:OPF-cost}
    \vskip -0.05in
\end{figure}

Under fixed load demand and generator configurations, we solve ACOPF for each generated scenario and record the resulting objective value, which reflects total generation cost and serves as a proxy for economic efficiency. Fig.~\ref{fig:OPF-cost} shows the ACOPF cost distributions for grids generated by the proposed method and the random-walk baseline, with the vertical dashed line indicating the reference cost of the original EU 36-bus system.

The random-walk baseline exhibits a broader and upward-shifted distribution, suggesting inefficient dispatch and distorted power-flow patterns induced by uncontrolled structural perturbations. In contrast, the proposed method produces a distribution tightly concentrated around the reference value, indicating preservation of economically meaningful dispatch structures. Notably, some generated grids achieve slightly lower ACOPF costs, reflecting structurally efficient yet physically plausible variations. This alignment is important for planning studies, as it demonstrates that the generated scenarios maintain realistic economic behavior under optimization rather than merely satisfying feasibility constraints.

\subsubsection{Load Stress and Shedding Analysis}
We further evaluate the stress resilience of the generated grids under increasing demand, motivated by high-load operating conditions or partial generation shortfalls in practice. Specifically, we scale the nominal load of the IEEE 14-bus system by a factor $\rho>1$ and assess the minimum fraction of total demand that must be shed to restore AC feasibility. For each generated grid, loads at high-demand buses are progressively curtailed until power-flow convergence is achieved, and the resulting shed fraction is recorded as a resilience metric.

\begin{figure}[h]
    \centering
    \vskip -0.1in
    \includegraphics[width=1\linewidth]{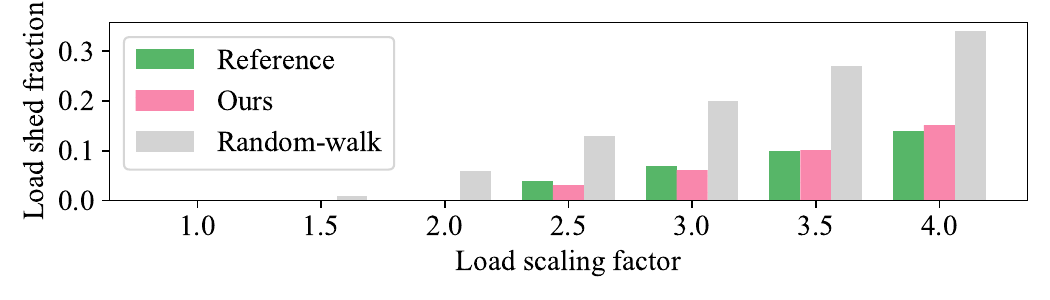}
    \vskip -0.15in
    \caption{Load-shedding fraction against load scaling factor $\rho$ in 14-bus system.}
    \label{fig:load_shed}
    \vskip -0.1in
\end{figure}

Fig.~\ref{fig:load_shed} reports the average load-shedding fraction as a function of $\rho$ for the reference grid, the random-walk baseline, and the diffusion-generated grids. The random-walk topologies begin requiring load shedding at relatively low stress levels ($\rho<2$), indicating limited redundancy and weakened transmission structure. In contrast, the diffusion-generated grids closely follow the reference system behavior up to $\rho \approx 3.5$, with only gradual shedding thereafter. These results demonstrate that the proposed method preserves structural characteristics critical to stress resilience, including effective power redistribution paths and generation–load connectivity, rather than merely satisfying nominal feasibility.

\vspace{-0.5em}
\section{Conclusion}
\label{sec:conclusion}

This paper presented a feasibility-aware framework for learning the AC-operable joint distribution of synthetic power-grid scenarios. By exploiting the natural hierarchy among network topology, branch electrical parameters, and operating conditions, the proposed framework decomposes high-dimensional scenario generation into computationally practical conditional stages while incorporating AC power-flow solvability and operational constraints directly into distribution learning. Numerical results on transmission and distribution benchmark systems demonstrate that the proposed method improves operational feasibility and contingency robustness, preserves realistic structural, electrical, and temporal characteristics, and supports downstream power-system applications. Future work will extend the framework to larger-scale power systems and richer operational scenarios involving renewable generation, distributed energy resources, and energy storage.
\bibliographystyle{IEEEtran}
\bibliography{ref}

\end{document}